%% file: paper.tex
\documentclass[]{style}

\input{preamble}

\usepackage[utf8]{inputenc} 
\usepackage[T1]{fontenc}    
\usepackage{graphicx}
\usepackage{amsmath}
\usepackage{amssymb}
\usepackage{url}            
\usepackage{booktabs}       
\usepackage{amsfonts}       
\usepackage{nicefrac}       
\usepackage{microtype}      
\usepackage{xcolor}         
\usepackage{multirow}
\usepackage{comment}
\usepackage{colortbl}
\usepackage{tocloft}
\usepackage{algorithm}
\usepackage{algorithmicx}
\usepackage{algpseudocode}
\usepackage{wrapfig}
\usepackage{tabularx}
\usepackage{textcomp}
\usepackage{stfloats}
\usepackage{verbatim}
\usepackage{titlesec}
\usepackage{adjustbox}
\usepackage{pifont}
\usepackage{tikz}
\usepackage{color}
\usepackage{subcaption}
\usepackage{soul}

\RequirePackage{xspace}
\makeatletter
\DeclareRobustCommand\onedot{\futurelet\@let@token\@onedot}
\def\@onedot{\ifx\@let@token.\else.\null\fi\xspace}

\makeatother

\definecolor{lightblue}{rgb}{0.66, 0.85, 0.95}
\hypersetup{
    colorlinks=true,
    linkcolor=red,
    citecolor=blue,
}
\definecolor{c2}{HTML}{FBD9BD}
\definecolor{c3}{HTML}{fe793d}
\definecolor{c4}{HTML}{eedeb0}
\definecolor{rouse}{rgb}{0.981,0.961,0.941}
\usepackage[most]{tcolorbox}
\tcbset{colback=blue!5!white, colframe=blue!20!white, boxrule=0pt, arc=2mm, left=1mm, right=1mm, top=1mm, bottom=1mm}

\definecolor{adptorange}{RGB}{248, 205, 172}
\definecolor{cmpblue}{RGB}{189, 215, 238}
\definecolor{cmpblue}{RGB}{189, 215, 238}

\definecolor{our_red}{RGB}{232,157,160}
\definecolor{our_blue}{RGB}{136,206,230}
\definecolor{our_orange}{RGB}{246,200,168}
\definecolor{our_green}{RGB}{178,211,164}

\definecolor{attn_code0}{RGB}{247,215,200}
\definecolor{attn_code1}{RGB}{238,169,139}
\definecolor{mlp_code0}{RGB}{204,201,221}
\definecolor{mlp_code1}{RGB}{102,95,153}

\definecolor{token_blue}{RGB}{84, 120, 140}

\usepackage{pifont}       
\usepackage{bbding}       
\usepackage{fontawesome}
\usepackage{xspace}

\usepackage{float}

\newlength\savewidth

\newcolumntype{x}[1]{>{\centering\arraybackslash}p{#1pt}}
\newcolumntype{y}[1]{>{\raggedright\arraybackslash}p{#1pt}}
\newcolumntype{z}[1]{>{\raggedleft\arraybackslash}p{#1pt}}

\renewcommand{\paragraph}[1]{\vspace{1mm}\noindent\textbf{#1}}

\usepackage{colortbl}
\usepackage{xcolor}
\usepackage{wrapfig}

\renewcommand{\paragraph}[1]{\vspace{1.25mm}\noindent\textbf{#1}}

\usepackage{listings}

\definecolor{codeblue}{rgb}{0.21, 0.49, 0.74}
\definecolor{codekw}{rgb}{0.35, 0.35, 0.75}
\lstdefinestyle{Pytorch}{
    language = Python,
    backgroundcolor = \color{white},
    basicstyle = \fontsize{9pt}{8pt}\selectfont\ttfamily\bfseries,
    columns = fullflexible,
    aboveskip=1pt,
    belowskip=1pt,
    breaklines = true,
    captionpos = b,
    commentstyle = \color{codeblue},
    keywordstyle = \color{codekw},
}

\definecolor{green}{HTML}{009000}
\definecolor{red}{HTML}{ea4335}

\title{The Pulse of Motion: Measuring Physical Frame Rate from Visual Dynamics}
\author[1]{Xiangbo Gao}
\author[1]{Mingyang Wu}
\author[1]{Siyuan Yang}
\author[1]{Jiongze Yu}
\author[1]{Pardis Taghavi}
\author[1]{Fangzhou Lin}
\author[1,2]{Zhengzhong Tu}

\affiliation[1]{Texas A\&M University}
\affiliation[2]{Visko Platform}

\abstract{
While recent generative video models have achieved remarkable visual realism and are being explored as world models, true physical simulation requires mastering both space and time. Current models can produce visually smooth kinematics, yet they lack a reliable internal motion pulse to ground these motions in a consistent, real-world time scale. This temporal ambiguity stems from the common practice of indiscriminately training on videos with vastly different real-world speeds, forcing them into standardized frame rates. This leads to what we term \textit{chronometric hallucination}: generated sequences exhibit ambiguous, unstable, and uncontrollable physical motion speeds. To address this, we propose Visual Chronometer, a predictor that recovers the Physical Frames Per Second (PhyFPS) directly from the visual dynamics of an input video. Trained via controlled temporal resampling, our method estimates the true temporal scale implied by the motion itself, bypassing unreliable metadata. To systematically quantify this issue, we establish two benchmarks, \textbf{\texttt{PhyFPS-Bench-Real}} and \textbf{\texttt{PhyFPS-Bench-Gen}}. Our evaluations reveal a harsh reality: state-of-the-art video generators suffer from severe PhyFPS misalignment and temporal instability. Finally, we demonstrate that applying PhyFPS corrections significantly improves the human-perceived naturalness of AI-generated videos.
\vspace{3mm}
\begin{quote}
\emph{``Not only do we measure the movement by the time, but also the time by the movement, because they define each other.''}\
--- Aristotle, \emph{Physics}
\end{quote}
\vspace{3mm}

}

\project{\url{https://xiangbogaobarry.github.io/Visual_Chronometer/}}
\date{\today}
\contact{Xiangbo Gao (\href{mailto:xiangbogaobarry@gmail.com}{xiangbogaobarry@gmail.com}), Zhengzhong Tu (\href{mailto:tzz@tamu.edu}{tzz@tamu.edu})}

\begin{document}
\thispagestyle{firstheader}
\maketitle
\pagestyle{plain}

\input{sec/intro}
\input{sec/related_works}
\input{sec/method}
\input{sec/exp}
\input{sec/con}

\newpage
{
\small
\bibliographystyle{IEEEtran}
\bibliography{main}
}


\end{document}

%% file: preamble.tex
\definecolor{lightblue}{rgb}{0.0, 0.6, 1.0}
\definecolor{darkgreen}{rgb}{0.0, 0.6, 0.0}
\definecolor{lightgreen}{rgb}{0.1, 0.8, 0.1}
\definecolor{lightred}{rgb}{0.7, 0.7, 0.7}
\definecolor{lightgreen}{rgb}{0, 0, 0}
\definecolor{lightlightgray}{rgb}{0.8, 0.8, 0.8}

\usepackage{titlesec}
\titlespacing*{\paragraph}
{0pt}{0em}{0.2em}  

\makeatletter
\renewcommand\paragraph{\@startsection{paragraph}{4}{\z@}%
  {3pt}
  {-0.5em}
  {\normalfont\normalsize\bfseries}} 
\makeatother

\usepackage{amsmath}  
\usepackage{upgreek}
\usepackage{array}    
\usepackage{multirow}
\usepackage{array}
\usepackage{algorithm}
\usepackage{makecell}
\usepackage{graphicx}
\usepackage{algorithm}
\usepackage{algpseudocode}
\usepackage{colortbl}
\usepackage{enumitem}
\usepackage{wrapfig}
\usepackage{mathrsfs}
\usepackage{pifont}
\usepackage{soul}
\usepackage{amssymb}
\usepackage{tcolorbox}
\usepackage{dirtytalk}
\usepackage{xspace}
\usepackage[utf8]{inputenc}

\usepackage{booktabs}

\usepackage{url}
\definecolor{blue}{rgb}{0.21,0.49,0.74}
\definecolor{red}{rgb}{0.8, 0.2, 0.2}
\definecolor{green}{rgb}{0, 0.5, 0}
\definecolor{yellow}{RGB}{218, 160, 109}
\definecolor{gray}{RGB}{155, 155, 155}

\crefname{section}{Sec.}{Secs.}
\Crefname{section}{Section}{Sections}
\Crefname{table}{Table}{Tables}
\crefname{table}{Tab.}{Tabs.}
\crefname{figure}{Fig.}{Figs.}
\Crefname{figure}{Figure}{Figures}
\crefname{appendix}{App.}{Apps.}
\Crefname{appendix}{Appendix}{Appendices}

%% file: sec/intro.tex
\section{Introduction}
\label{sec:intro}

While modern generative video models excel at spatial realism—producing photorealistic textures, complex geometry, and coherent layouts~\cite{wan2025wan,hacohen2026ltx,jiang2025vace,burgert2025motionv2v,gao2026pisco, wu2026consid}—an increasing number aspire to go further and act as physical world models~\cite{ali2025world, team2026advancing}. However, faithfully simulating the physical world requires an intricate mastery of both space and time; physical motion is governed by a strict relationship between spatial displacement and elapsed time, yet today's video generation pipelines often lack a stable pulse of motion to track this. Consequently, while modern generators can produce visually fluid kinematics, these motions are rarely grounded in a consistent, real-world time scale.

\begin{figure}
    \centering
    \includegraphics[width=1\linewidth]{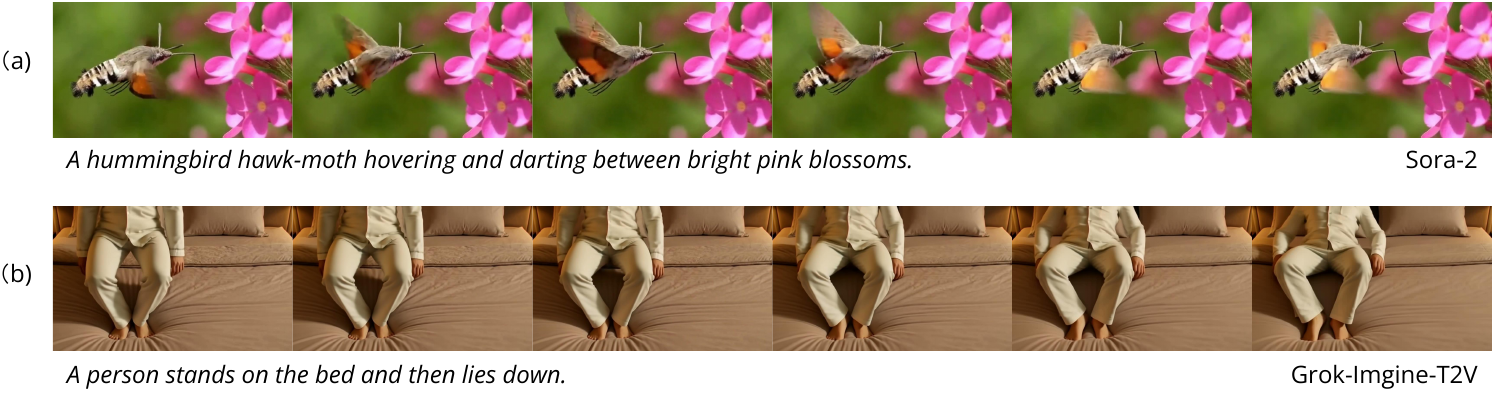}
    \vspace{-2mm}
    \caption{\textbf{Visualization of Chronometric Hallucination.} Current video generators sometimes fail to ground their outputs in a consistent physical time scale, even when no speed-manipulating keywords (e.g., ``slow motion'') are prompted. \textbf{(a)} A hummingbird hawk-moth is rendered in extreme slow-motion, despite its naturally high wing-beat frequency. \textbf{(b)} A person falls onto a bed at a velocity significantly slower than standard gravity. These instances illustrate \textbf{Chronometric Hallucination}: a prevalent failure mode where generated motions exhibit an ambiguous, unstable, and uncontrollable physical time scale.}
    \label{fig:chronometric_hallucination}
    \vspace{-4mm}
\end{figure}

Much of this temporal ambiguity stems from the agnostic treatment of time during the training of modern video models~\cite{wu2025densedpo}. Internet-scale video datasets are mixtures of varying capture and editing regimes, encompassing standard-rate footage, extreme slow-motion, and accelerated time-lapses. During training, models are typically blind to these inherent physical speeds; a time-lapse and a slow-motion video might be fed into the network identically. This lack of time-scale awareness severs the correspondence between a discrete frame step and the real-world time elapsed. As a result, models learn to generate plausible frame-to-frame transitions, but the underlying physical speed of the generated motion becomes ambiguous, unstable, and impossible to explicitly control. We refer to this prevalent failure mode as \textbf{Chronometric Hallucination} (see ~\Cref{fig:chronometric_hallucination}).

Aristotle once observed that \say{\emph{not only do we measure the movement by the time, but also the time by the movement, because they define each other.}} Operationalizing this ancient principle, we introduce \textbf{Visual Chronometer}, a predictor designed to alleviate chronometric hallucination by recovering this intrinsic motion pulse, formalized as \textbf{Physical Frames Per Second (PhyFPS)}, directly from visual dynamics. We distinguish the inherent PhyFPS from the nominal metadata (meta FPS) by defining PhyFPS as the true frame rate that aligns with the real-world passage of time. Through controlled temporal resampling, we supervise the model to learn these motion-grounded dynamics, bypassing the often unreliable metadata.

We evaluate Visual Chronometer across multiple dimensions. First, to validate the accuracy of our method, we introduce \textbf{\texttt{PhyFPS-Bench-Real}}, comprising real-world videos where the true PhyFPS often diverges from the meta FPS due to complex speed variations. Second, we establish \textbf{\texttt{PhyFPS-Bench-Gen}} to systematically audit state-of-the-art video generators along three complementary axes: (i) the alignment between meta FPS and actual PhyFPS, (ii) intra-video stability (the consistency of PhyFPS across sliding windows within a single clip), and (iii) inter-video stability across different outputs from the same model configuration.

Our extensive measurements reveal a harsh reality: even strong generators exhibit substantial PhyFPS misalignment, alongside significant intra- and inter-video temporal jitter. Without a grounded physical time scale, these models fail to provide the reliable simulation necessary for true world modeling. Furthermore, we demonstrate that applying PhyFPS-guided post-corrections to generated videos substantially improves human-perceived naturalness, as validated by our user study. Finally, we evaluate strong Vision-Language Models (VLMs) as potential PhyFPS judges, finding them vastly unreliable for this specialized task, thereby underscoring the necessity of our dedicated Visual Chronometer. Our contributions are summarized as follows:
\begin{itemize}[leftmargin=1.2em, itemsep=0.2em, nosep]
\item We identify and define the phenomenon of \textbf{chronometric hallucination} in modern video generators, and formalize \textbf{Physical Frames Per Second (PhyFPS)} as a temporal scale distinct from nominal meta FPS.

\item We propose \textbf{Visual Chronometer}, a robust predictor that recovers PhyFPS directly from raw frames by learning motion-grounded dynamics through controlled temporal resampling.

\item We introduce \texttt{PhyFPS-Bench-Gen} to audit state-of-the-art generators, revealing severe time-scale misalignment in modern video generators. We further show that PhyFPS-guided post-correction significantly enhances human-perceived temporal naturalness.
    
\item Through \texttt{PhyFPS-Bench-Real}, we demonstrate our model's precision in predicting Physical FPS. Our analysis also reveals that the state-of-the-art VLMs are unreliable temporal judges, underscoring the necessity of a dedicated Visual Chronometer.

\end{itemize}

%% file: sec/related_works.tex
\section{Related Works}

\subsection{Video Generation and the Quest for World Models}
\label{sec:rw_world_models}

Modern video generative models, spanning large-scale diffusion and autoregressive architectures, have achieved unprecedented perceptual quality and semantic coherence 
\cite{InfinityStar,wan2025wan,hacohen2026ltx,hacohen2024ltx,yang2024cogvideox,hong2022cogvideo,kong2024hunyuanvideo,elmoghany2026infinitystory,wu2026consid,yu2026sparkvsr}. 
To capture dynamics, these systems employ sophisticated temporal modeling mechanisms, such as 3D spatiotemporal operators~\cite{tran2015learning, vaswani2017attention}, causal attention blocks~\cite{ali2025world, InfinityStar}, and temporal latent spaces~\cite{tong2022videomae, xing2024large}. As these architectures scale, they are increasingly framed as ``world models'' capable of simulating physical environments~\cite{kang2024far, qin2024worldsimbench, ding2025understanding, wang2026mechanistic, wang2025generative}.
However, while prior works focus heavily on optimizing frame-to-frame kinematic smoothness and spatial layout, the actual physical time scale of the depicted motion is rarely encoded or supervised~\cite{yuan2025newtongen, gao2025seeing}; instead, models rely entirely on the nominal frame rate (meta FPS) provided by the dataset container. 
Because these advanced generative mechanisms do not explicitly ground their temporal learning in real-world physics, they remain highly vulnerable to chronometric hallucination---producing motions that look perceptually smooth but lack a consistent physical speed.
We argue that one cannot fix a physical flaw without first being able to measure it. Thus, we complement these generative advancements by developing the first dedicated tool to audit this structural blind spot. By explicitly defining and predicting the intrinsic Physical FPS (PhyFPS), we provide the necessary metric and benchmark to evaluate time-scale calibration in world models.

\subsection{Visual Perception of Time and Dynamics}
\label{sec:rw_timescale_speed}

Our methodology draws inspiration from a long-standing line of computer vision research aimed at understanding time and speed from visual cues. 
Early efforts in this domain focused on domain-specific heuristics, such as detecting slow-motion replays in sports broadcasts \cite{wang2004generic,chen2015novel,kiani2012effective}. 
More recently, self-supervised approaches like SpeedNet \cite{benaim2020speednet} demonstrated that neural networks can discriminate between normal-rate and artificially sped-up clips. In a parallel vein, research on the ``arrow of time'' explores whether models can recognize the forward or backward directionality of video playback \cite{pickup2014seeing,wei2018learning}. 
Furthermore, semantic hyperlapse and time-remapping techniques actively manipulate temporal sampling to summarize videos \cite{bennett2007computational,petrovic2005adaptive,zhou2014time,lan2018ffnet,silva2018weighted,da2019semantic}, proving that visual dynamics naturally dictate the perceived flow of time.
However, these existing perception models typically frame time as a binary classification problem (e.g., faster vs.\ slower, forward vs.\ backward). They do not aim to recover a high-precision physical metric. 
In contrast, \textbf{Visual Chronometer} frames time-scale perception as an absolute continuous regression problem, directly predicting PhyFPS from frame sequences to audit generative models without relying on corrupted metadata.



\subsection{Benchmarking Temporal and Physical Fidelity}
\label{sec:rw_benchmark_metrics}

Evaluating video generation has traditionally been dominated by perceptual quality and semantic fidelity metrics. 
Standard protocols rely on frame-level similarity (PSNR \cite{jahne2005digital}, SSIM \cite{wang2004image}, LPIPS \cite{zhang2018unreasonable}), no-reference perceptual quality predictors for user-generated and variable-frame-rate videos such as RAPIQUE \cite{tu2021rapique} and FAVER \cite{zheng2024faver}, and distribution-level feature matching, most notably the Fréchet Video Distance (FVD) \cite{skorokhodov2022stylegan}. 
Recognizing the limitations of monolithic metrics, recent comprehensive suites like VBench \cite{huang2024vbench,zheng2025vbench,huang2025vbench++} and WorldScore \cite{duan2025worldscore} have introduced multi-dimensional evaluations, including physics-adjacent axes such as temporal consistency and action alignment. 
Nevertheless, these benchmarks primarily evaluate whether the motion ``looks natural'' rather than measuring the exact temporal speed governing the scene. Time-scale fidelity---specifically, whether a video strictly adheres to a stable physical frame rate throughout its duration---remains entirely unmeasured. 
Our introduced benchmarks, \texttt{PhyFPS-Bench-Real} and \texttt{PhyFPS-Bench-Gen}, fill this critical void. By shifting the evaluation paradigm from perceptual smoothness to chronometric measurement, we provide the first quantitative audit of intra-video and inter-video time-scale stability in generative world models.

%% file: sec/method.tex
\section{Data Preparation}
\label{sec:method} 

\subsection{Data Collection}
\label{sec:data_collection}

To train \textbf{Visual Chronometer} to accurately predict the \textbf{Physical Frames Per Second (PhyFPS)}, we require a training dataset with verified, ground-truth temporal labels. A model trained on data suffering from chronometric hallucination inherently cannot serve as a reliable temporal measurement tool. Therefore, we curate a dataset exclusively from video sources where the nominal metadata frame rate perfectly aligns with the real-world physical sampling rate (i.e., meta FPS = PhyFPS), strictly excluding videos with ambiguous post-hoc time-scale editing. We aggregate our high-fidelity source data from the following categories:

\begin{itemize}[leftmargin=1.2em, itemsep=0.2em, nosep]
\item \textbf{High-Frame-Rate Academic Datasets:} We utilize high-speed benchmarks, including Adobe240~\cite{su2017deep} and BVI-VFI~\cite{danier2023bvi} (up to 120~Hz), typically used for precise temporal analysis and frame interpolation.

\item \textbf{Raw Broadcast Sequences:} Uncompressed 4K YUV footage from UVG~\cite{mercat2020uvg} (50/120 FPS) is included; its raw pipeline minimizes the risk of hidden temporal remapping.

\item \textbf{Sensor-Synchronized Autonomous Data:} Datasets from NVIDIA and Honda~\cite{ramanishka2018toward} provide cross-sensor alignment (Camera/LiDAR/IMU), where strict synchronization guarantees physical time-scale integrity.

\item \textbf{Physics-Grounded Human Motion:} Human-centric sequences~\cite{mehta2017monocular} are incorporated to leverage motion captured specifically for biomechanical and dynamic realism.

\item \textbf{Verified In-House Data:} We supplement the public datasets with an internal collection captured under strictly controlled settings with verified frame-rate metadata.
\end{itemize}

\subsection{Data Preprocessing and Augmentation}
\label{sec:data_preprocessing}

To force the model to learn intrinsic visual dynamics rather than relying on semantic content priors, we expand our training distribution by synthetically generating a diverse array of PhyFPS variants from the source videos. We first temporally upsample all source videos to a high-frequency base rate of 240 FPS using a state-of-the-art frame interpolation model (RIFE)~\cite{huang2022rife}. 

Let this high-rate video be $I^{H}$ at a frequency $F_H=240$ FPS. For a target lower frame rate $F_L$, we define the downsampling ratio as $N=F_H/F_L$. We then synthesize low-rate videos ($I^{L}$) using three distinct strategies (illustrated in Fig.~\ref{fig:datademo}), each designed to model specific real-world camera mechanics:

\begin{figure*}[t]
    \centering
    \vspace{-2mm}
    \begin{minipage}[b]{0.69\linewidth}
        \centering
        \includegraphics[width=0.95\linewidth]{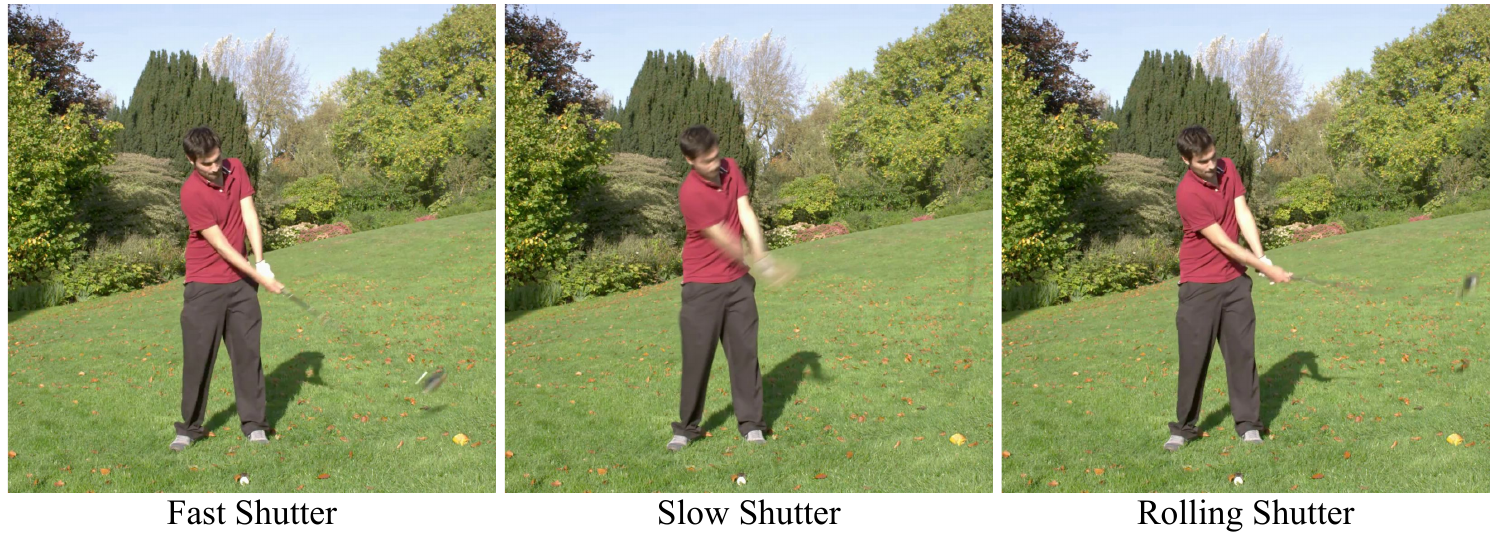}
        \caption{\textbf{Physics-Grounded Temporal Augmentation.} We synthesize diverse low-rate videos from high-frequency source data (240 FPS) to simulate real-world camera mechanics: Sharp Capture, Motion Blur, and Rolling Shutter.}
        \label{fig:datademo}
    \end{minipage}
    \hfill
    \begin{minipage}[b]{0.28\linewidth}
        \centering
        \includegraphics[width=0.9\linewidth]{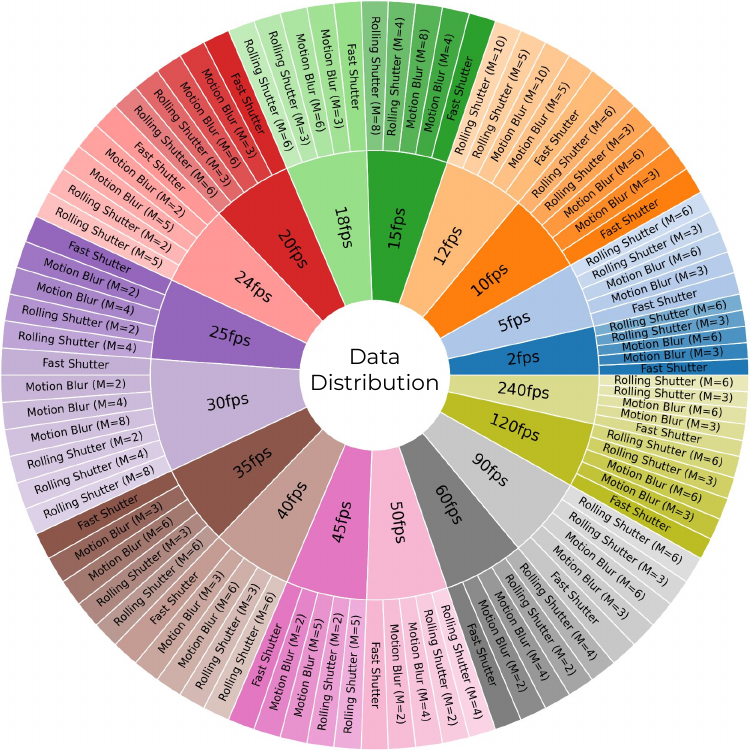}
        \caption{Dataset distribution across 18 target Physical Frame Rates.}
        \label{fig:data_distribution}
    \end{minipage}
    \vspace{-2mm}
\end{figure*}

\paragraph{(1) Sharp Capture (Fast Shutter):} 
To simulate cameras operating with a very fast shutter speed (which minimizes motion blur), we uniformly subsample the high-rate sequence by setting $I^{L}_{k}=I^{H}_{\lfloor kN \rfloor}$, where $I^{L}_{k}$ denotes the $k$-th frame of the synthesized low-rate video and $\lfloor kN \rfloor$ is the corresponding discrete frame index in the high-rate source. This isolates pure spatial displacement over time, preserving sharp object boundaries but often resulting in the naturally aliased motion (stutter) typically seen in sports or action footage.

\paragraph{(2) Motion Blur (Variable Exposure):} 
Real-world cameras integrate light over an exposure window, resulting in motion blur that provides strong visual cues about object velocity. To mimic this exposure integration, we synthesize each low-rate frame by averaging a temporal window of high-rate frames: $I^{L}_{k}=\frac{1}{M}\sum_{i=0}^{M-1} I^{H}_{\lfloor kN \rfloor + i}$, where $M$ is the exposure window length. We simulate long, medium, and short effective exposures by setting $M \in \{N, N/2, N/4\}$.

\paragraph{(3) Synthetic Rolling Shutter}
Fast-moving objects captured by modern CMOS sensors frequently exhibit rolling shutter distortions~\cite{liang2005rolling} because sensor rows or columns are read sequentially rather than instantaneously. We simulate this intra-frame temporal distortion by partitioning the target frame's spatial dimension (e.g., width $W$) into progressive bands. A pixel at column $x$ is sampled from the high-rate sequence at a progressively shifted time index: $\lfloor kN \rfloor + \lfloor M \cdot x / W \rfloor$. By varying the readout duration $M \in \{N, N/2, N/4\}$, we ensure the predictor is robust to these common spatiotemporal artifacts.

\paragraph{Final Dataset Composition.}
As summarized in Fig.~\ref{fig:data_distribution}, alongside these synthetically augmented variants, we retain the original source videos at their native capture rates to preserve raw sensor statistics. e generate training data across 18 Physical Frame Rates, yielding a comprehensive dataset of 465,535 video clips, uniformly standardized to a length of 128 frames to ensure balanced representation across different time scales.

\section{Visual Chronometer}

\subsection{Model Architecture}
\label{sec:architecture}

\paragraph{Backbone and Regression Head.}
We adopt VideoVAE+~\cite{xing2024large} as the foundational video encoder to extract compact spatiotemporal latent representations. Given an input clip of $T$ frames $\mathbf{V}=\{I_t\}_{t=1}^{T}$, the backbone produces a sequence of latent tokens $\mathbf{Z} = \text{Enc}(\mathbf{V})$. 

Instead of relying on conventional spatial pooling, we attach a lightweight, attention-based prediction head to aggregate temporal features into a clip-level representation. Specifically, we project the latent tokens into a hidden dimension and introduce a learnable query embedding that cross-attends to the token sequence. This query-based pooling mechanism effectively decouples the regression head from the input frame count, enabling \textbf{Visual Chronometer} to process videos of arbitrary lengths. Finally, a Multi-Layer Perceptron (MLP) maps the aggregated feature vector to a single scalar $\hat{s} \in \mathbb{R}$, which represents the predicted logarithmic frame rate, $\log(\text{PhyFPS})$. We predict the logarithmic value rather than the absolute frequency to stabilize optimization across an exponentially wide range of time scales and to penalize relative, rather than absolute, errors.

\subsection{Training Objective}
\label{sec:loss}

Let the ground-truth PhyFPS be $y$, with its log-space target defined as $s=\log y$. The model outputs the prediction $\hat{s}=\log \hat{y}$. We optimize the model using a Mean Squared Error (MSE) in the logarithmic space:
\begin{equation}
\label{eq:log_mse}
    \mathcal{L}_{\log} \;=\; \frac{1}{n}\sum_{i=1}^{n} \left( \log y_i - \hat{s}_i \right)^2,
\end{equation}
where $n$ is the batch size. Because the target PhyFPS values in our dataset are strictly positive ($y_i \ge 2$), the logarithmic transformation is intrinsically well-defined. Therefore, we deliberately omit the standard offset term ($+1$) typically found in traditional Mean Squared Logarithmic Error (MSLE) formulations, allowing the loss to strictly reflect the true proportional scaling of time.

\subsection{Model Training Details}
\label{sec:training}

To train the Visual Chronometer, we extract clips from the dataset using a sliding window. During training, clips are sampled with a maximum temporal footprint of $T=32$ frames. To ensure robust performance across different deployment scenarios, we train two variants of the model targeting different operational regimes. The \textbf{VC-Wide} model is trained to predict across 18 distinct frame rates spanning from extreme slow-motion to high-speed capture: $\text{PhyFPS} \in \{2, 5, 10, 12, 15, 18, 20, 24, 25, 30, 35, 40, 45, 50, 60, 90, 120, 240\}$. The \textbf{VC-Common} model focuses specifically on the most prevalent consumer and web video formats, narrowing the output space to $\text{PhyFPS} \in \{12, 15, 18, 20, 24, 25, 30, 35, 40, 45, 50, 60\}$. 

Both models are trained end-to-end, fine-tuning the VideoVAE+ backbone jointly with the attention-based prediction head. Optimization is performed using the Adam optimizer with a learning rate of $1 \times 10^{-5}$ for 125,000 iterations. We execute the training on a single computing node equipped with four NVIDIA RTX A6000 GPUs, utilizing a global batch size of 32.

%% file: sec/exp.tex
\section{Experiments}
\label{sec:experiments}

In this section, we conduct three sets of experiments to validate the \textbf{Visual Chronometer} and demonstrate its utility in addressing chronometric hallucination, as well as enabling physics-grounded data preprocessing and video post-processing. 
First, we introduce \textbf{\texttt{PhyFPS-Bench-Gen}} to audit existing open- and closed-source video generative models by measuring their Meta-vs-PhyFPS alignment and temporal stability. Second, we build \textbf{\texttt{PhyFPS-Bench-Real}} to evaluate the prediction accuracy of our model against reliable ground-truth labels. Third, we compare our specialized predictor against strong Vision-Language Models (VLMs), demonstrating that general-purpose foundation models are not yet capable of reliable PhyFPS prediction.

\subsection{Auditing Generative World Models}
\label{sec:gen_fps_bench}

\paragraph{\texttt{PhyFPS-Bench-Gen}.}
We introduce \textbf{\texttt{PhyFPS-Bench-Gen}}, a benchmark designed to quantitatively audit the time-scale alignment of video generative models using our Visual Chronometer. We evaluate a diverse spectrum of leading generators. For open-source models, we assess the Wan series~\cite{wan2025wan} (Wan2.1-1.3B, Wan2.1-14B, Wan2.2-5B, Wan2.2-14B), the LTX series~\cite{hacohen2024ltx,hacohen2026ltx} (LTX-Video, LTX-2), the CogVideoX series~\cite{yang2024cogvideox} (CogVideoX-2B, CogVideoX-5B), HunyuanVideo~\cite{kong2024hunyuanvideo}, and the autoregressive model InfinityStar~\cite{liu2025infinitystar}. For closed-source models, we evaluate Veo-3.1-Fast~\cite{deepmind_veo3_2025}, Sora-2~\cite{openai_sora_2024}, Grok-Imagine-T2V~\cite{xai_grok_imagine_2026}, Kling-o3~\cite{klingai_omninew_2025}, Seedance-1.0-Lite~\cite{bytedance2025seed16flash}, and Seedance-1.5-Pro~\cite{bytedance2025seed16flash}.

\paragraph{Benchmark Prompts.}
To ensure robust evaluation, we design 100 text-to-video prompts covering diverse content and motion patterns, strictly avoiding explicit speed-manipulation keywords (e.g., \texttt{slow motion}, \texttt{time-lapse}, \texttt{speed up}). To guarantee that PhyFPS is observable, every prompt mandates at least one clearly dynamic instance, excluding purely static scenes. Prompt diversity is balanced across five axes: (i) \textbf{primary entity} (human, animal, vehicle, and nature), (ii) \textbf{motion type} (articulated, rigid-body, fluid, and multi-agent), (iii) \textbf{camera behavior} (static, pan, and tracking), (iv) \textbf{environmental effects} (rain, fire, and wind), and (v) \textbf{scene context} (indoor, urban, and nature). All models operate under default settings, extracting the nominal saved FPS ($F_{\text{meta}}$) from official documentation or output metadata, where $F_{\text{meta},c}$ denotes the container FPS of the source video corresponding to clip $c$.

\paragraph{PhyFPS Estimation and Metrics.}
For all audits on generated videos, we employ the \textbf{VC-Common} predictor. For each video $v \in \{1,\dots,V\}$, we extract $C_v$ overlapping clips of $T{=}32$ frames with stride $s{=}4$. Let $\hat{f}_{v,c}$ denote the predicted PhyFPS for clip $c$. The video-level PhyFPS ($\bar{f}_v$) and the overall model-level PhyFPS ($\hat{F}$) are computed as:
\begin{equation}
\label{eq:avg_fps}
    \bar{f}_v \;=\; \frac{1}{C_v}\sum_{c=1}^{C_v} \hat{f}_{v,c}, 
    \qquad
    \hat{F} \;=\; \frac{1}{V}\sum_{v=1}^{V} \bar{f}_v.
\end{equation}

We evaluate each generator along three critical dimensions. \textbf{(1) Meta-vs-PhyFPS Alignment} measures how well the nominal container rate $F_{\text{meta}}$ matches the predicted intrinsic speed. We report both the \textbf{Avg. Error (FPS)} and the \textbf{Pct. Error (\%)}:
\begin{equation}
\label{eq:align_metrics}
\begin{aligned}
\text{Avg. Error} 
&= \frac{1}{V} \sum_{v=1}^{V} \frac{1}{C_v} \sum_{c=1}^{C_v}
\left| \hat{f}_{v,c} - F_{\text{meta},c} \right|, \quad
\text{Pct. Error} 
&= \frac{100}{V} \sum_{v=1}^{V} \frac{1}{C_v} \sum_{c=1}^{C_v}
\frac{\left| \hat{f}_{v,c} - F_{\text{meta},c} \right|}{F_{\text{meta},c}}.
\end{aligned}
\end{equation}
\textbf{(2) Inter-video Consistency} and \textbf{(3) Intra-video Consistency} evaluate temporal stability across different prompts and within a single continuous video, respectively. Both utilize the coefficient of variation (CV):
\begin{equation}
\label{eq:cv_metrics}
    \text{Inter CV} = \frac{\mathrm{Std}\!\left(\{\bar{f}_v\}_{v=1}^{V}\right)}{\mathrm{Mean}\!\left(\{\bar{f}_v\}_{v=1}^{V}\right)}, \quad \text{Intra CV} = \frac{1}{V}\sum_{v=1}^{V} \frac{\mathrm{Std}\!\left(\{\hat{f}_{v,c}\}_{c=1}^{C_v}\right)}{\mathrm{Mean}\!\left(\{\hat{f}_{v,c}\}_{c=1}^{C_v}\right)}.
\end{equation}

\begin{table*}[t]
\centering
\scriptsize
\setlength{\tabcolsep}{8pt}
\vspace{-2mm}
\caption{\textbf{Quantitative Audit of Generative Models.} \texttt{PhyFPS-Bench-Gen} results evaluating time-scale fidelity. \textcolor{blue}{Blue} and \textcolor{red}{red} shaded cells indicate the best and second-best performance within each group (open vs.\ closed source).}
\label{tab:fps_metrics}
\renewcommand{\arraystretch}{1}
\resizebox{\textwidth}{!}{%
\begin{tabular}{lcccccc}
\toprule
Model & Meta FPS & PhyFPS & Avg. Error $\downarrow$ & Pct. Error(\%)$\downarrow$ & Intra CV $\downarrow$ & Inter CV $\downarrow$ \\
\midrule
\multicolumn{7}{c}{\textit{Open-sourced Models}} \\
\midrule
CogVideoX-2B & 24 & 33.64 & 12.46 & 52 & 0.11 & 0.46 \\
CogVideoX-5B & 24 & 38.26 & 17.96 & 75 & 0.12 & 0.52 \\
HunyuanVideo & 24 & 35.89 & 13.82 & 58 & 0.12 & 0.36 \\
Wan2.1-T2V-1.3B & 24 & 26.28 & \cellcolor{blue!15}7.54 & \cellcolor{blue!15}31 & \cellcolor{red!15}0.11 & 0.38 \\
Wan2.1-T2V-14B & 24 & 32.37 & 10.87 & 45 & 0.14 & 0.36 \\
Wan2.2-T2V-A14B & 24 & 31.52 & \cellcolor{red!15}10.74 & \cellcolor{red!15}45 & 0.12 & 0.38 \\
Wan2.2-TI2V-5B & 24 & 32.81 & 11.63 & 48 & 0.15 & 0.38 \\
InfinityStar (5s) & 16 & 34.41 & 18.46 & 115 & 0.11 & 0.38 \\
InfinityStar (10s) & 16 & 36.15 & 20.19 & 126 & 0.16 & 0.36 \\
LTX-Video & 24 & 46.52 & 23.67 & 99 & \cellcolor{blue!15}0.10 & \cellcolor{blue!15}0.33 \\
LTX-2 & 25 & 39.77 & 15.70 & 63 & 0.13 & \cellcolor{red!15}0.34 \\
\midrule
\multicolumn{7}{c}{\textit{Closed-sourced Models}} \\
\midrule
Seedance-1.0-Lite & 24 & 28.60 & \cellcolor{blue!15}8.31 & \cellcolor{red!15}35 & 0.15 & 0.37 \\
Seedance-1.5-Pro & 24 & 33.69 & 10.67 & 44 & 0.16 & \cellcolor{blue!15}0.25 \\
Sora-2 & 30 & 36.21 & \cellcolor{red!15}8.40 & \cellcolor{blue!15}28 & \cellcolor{blue!15}0.13 & 0.29 \\
Grok-Imagine-T2V & 24 & 36.97 & 13.97 & 58 & 0.16 & \cellcolor{red!15}0.28 \\
Kling-o3 & 24 & 30.04 & 9.10 & 38 & \cellcolor{red!15}0.15 & 0.34 \\
Veo-3.1-Fast & 24 & 35.83 & 13.62 & 57 & 0.17 & 0.33 \\
\bottomrule
\end{tabular}%
}
\vspace{-2mm}
\end{table*}

\paragraph{Audit Results.}
\Cref{tab:fps_metrics} details the results of the \texttt{PhyFPS-Bench-Gen} audit. We observe a pervasive Meta-vs-PhyFPS mismatch across the majority of generators; despite outputs being stored at a fixed nominal meta FPS, the intrinsic visual speeds vary wildly. Notably, the Wan series models exhibit a relatively high adherence to their suggested meta FPS, achieving comparatively low average and percentage errors. To evaluate temporal consistency, we measure Inter CV and Intra CV, which represent the fluctuation of PhyFPS across different generated videos and the stability of PhyFPS across different time segments within a single video, respectively. The LTX-Video and LTX-2 models demonstrate strong performance on these stability metrics. This suggests their temporal representations are internally consistent, and the high absolute errors may primarily stem from inaccurate meta FPS metadata rather than structural chronometric hallucination. For instance, LTX-Video's outputs might simply need their meta FPS adjusted from 24 to around 46.5 to achieve high time-scale fidelity. 

Overall, closed-source models slightly outperform open-source counterparts in terms of absolute accuracy, maintaining average errors below 14 FPS and percentage errors under 60\%. This indicates that commercial models may employ more carefully designed strategies for selecting meta FPS. However, despite these advantages in global alignment, the Intra and Inter CV scores of closed-source models are not significantly better than those of open-source models. This reveals that even heavily optimized, industry-scale generators still struggle with PhyFPS stability. It appears researchers predominantly prioritize visual fidelity and kinematic smoothness, inadvertently neglecting strict physical time-scale adherence. This lack of reliable temporal grounding poses a significant challenge for leveraging current video generative models as accurate world models. Finally, we observe a consistent trend where the predicted PhyFPS is generally higher than the assigned Meta FPS across almost all videos. According to the Visual Chronometer, most generated videos should be played back at a higher meta FPS, or directly at their intrinsic PhyFPS. This finding aligns with the widely recognized phenomenon that current generative models tend to produce ``slow but smooth'' videos~\cite{wu2025densedpo}.

\paragraph{User Study: Perceptual Validation via Video Post-processing.}
To demonstrate the practical utility of our method and confirm that mathematical PhyFPS accuracy translates to improved human perception, we conduct a user study treating the predictor as a post-processing tool. Using \textbf{VC-Common}, we predict the clip-level PhyFPS for generated videos. We then present users with three variants of the same sequence. The first variant is the \textbf{Original} video, representing the untouched output directly from the generative model. The second variant, \textbf{Pred}, serves as a globally corrected version; here, we uniformly re-time the entire video to match its average predicted PhyFPS. The third variant, \textbf{Pred Dyn}, applies a dynamic local correction, where each distinct temporal segment within the video is independently re-timed based on its specific, clip-level PhyFPS prediction.

\begin{wrapfigure}{r}{0.49\columnwidth}
    \centering
    \vspace{-7mm}
    \includegraphics[width=\linewidth]{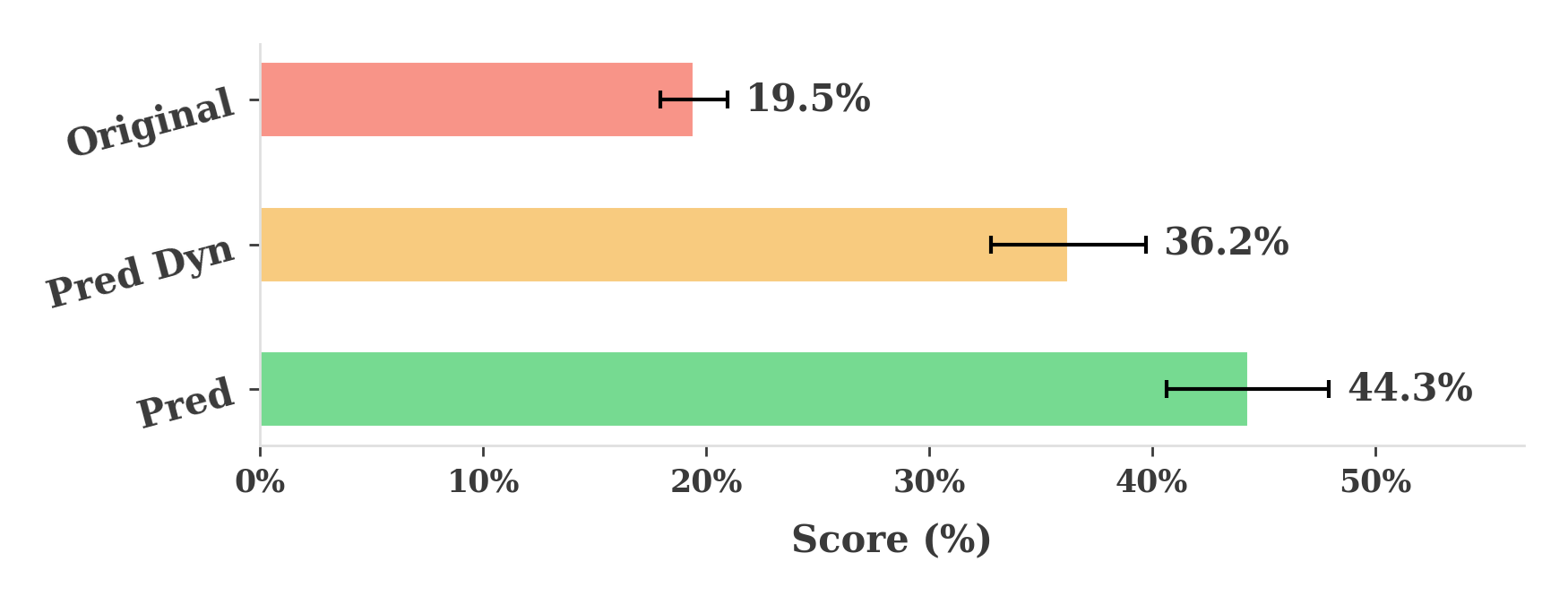}
    \caption{\textbf{Human Perceptual Preference on Temporal Naturalness.} Bradley-Terry scores comparing the original generated videos against our post-processed variants. Both the global average correction (\textbf{Pred}) and dynamic local correction (\textbf{Pred Dyn}) are strongly preferred over the hallucinated original outputs, with 90\% confidence intervals indicating statistical significance.}
    \label{fig:overall_bt}
    \vspace{-4mm}
\end{wrapfigure}

We collected 1,490 pairwise comparisons from over 15 participants. Utilizing the Bradley--Terry model~\cite{bradley1952rank}, we estimated the relative preference strength for each variant, computing 90\% confidence intervals via bootstrapping (\Cref{fig:overall_bt}). The results reveal that both post-processed variants significantly outperform the hallucinated original outputs (19.0\%). Interestingly, the global correction (Pred, 44.2\%) is preferred over the dynamic local correction (Pred Dyn, 36.9\%). We hypothesize that while dynamic correction perfectly aligns local clips to their intrinsic PhyFPS, varying the playback frame rate within a single short sequence may introduce perceptual inconsistencies or jitter. Conversely, applying a constant, averaged Physical Frame Rate (Pred) remains visually smoother and more natural to human observers. Ultimately, these findings definitively highlight the value of physics-grounded post-processing.

\subsection{Validating the Visual Chronometer}
\label{sec:pred_fps_bench}

To establish the reliability of our measurement tool, we evaluate its prediction accuracy on the \texttt{PhyFPS-Bench-Real} test set (comprising 4,000 verified clips partitioned from our dataset). Crucially, to ensure that the Visual Chronometer learns intrinsic physical time scales rather than overfitting to dataset-specific biases, we enforce a strict cross-source split; the training, validation, and test sets are derived from entirely disjoint video sources. Given the ground-truth PhyFPS $y_i$ and predicted PhyFPS $\hat{y}_i$ across $n$ test samples, we report the Mean Absolute Error (MAE) and Mean Absolute Percentage Error (MAPE) to capture both absolute deviations and proportional accuracy:
\begin{equation}
\label{eq:mae_mape}
    \text{MAE} \;=\; \frac{1}{n}\sum_{i=1}^{n}\left|y_i - \hat{y}_i\right|, 
    \qquad
    \text{MAPE} \;=\; \frac{100}{n}\sum_{i=1}^{n}\frac{\left|y_i - \hat{y}_i\right|}{y_i}.
\end{equation}
Given the rapid advancements in Vision-Language Models (VLMs), it is tempting to deploy them as out-of-the-box evaluators for physical scene dynamics. To rigorously test this hypothesis, we establish a comprehensive baseline using state-of-the-art VLMs, including Gemini-3.1-Pro~\cite{deepmind_gemini_pro_2025}, Gemini-3-Flash~\cite{deepmind_gemini_pro_2025}, Seed-1.6 and Seed-1.6-Flash~\cite{bytedance2025seed16flash}, as well as Qwen3.5+ and Qwen3.5-397B~\cite{yang2025qwen3}.

\begin{wraptable}{r}{0.48\textwidth}
\centering
\scriptsize
\setlength{\tabcolsep}{3pt}
\vspace{-4mm}
\caption{\textbf{Predictor Accuracy \& VLM Baseline Comparison.} Evaluating the Visual Chronometer (Ours) against state-of-the-art Vision-Language Models on \texttt{PhyFPS-Bench-Real}. The average ground-truth PhyFPS across the test set is 38.81. \textcolor{blue}{Blue} and \textcolor{red}{red} shaded cells indicate the best and second-best performance.}
\label{tab:fps_comparison}
\resizebox{\linewidth}{!}{%
\begin{tabular}{lccc}
\toprule
Model & Avg Pred & MAE $\downarrow$ & MAPE(\%)$\downarrow$ \\
\midrule
\multicolumn{4}{c}{\textit{Ours}} \\
\midrule
VC-Common      & 39.20 & \cellcolor{blue!15}3.46  & \cellcolor{blue!15}9 \\
VC-Wide        & 45.48 & \cellcolor{red!15}7.76  & \cellcolor{red!15}21 \\
\midrule
\multicolumn{4}{c}{\textit{Video-based VLM}} \\
\midrule
Gemini-3.1-Pro & 31.00 & 21.67 & 43 \\
Gemini-3-Flash & 26.60 & 23.40 & 47 \\
Seed-1.6       & 29.60 & 20.40 & 41 \\
Seed-1.6-Flash & 30.00 & 20.00 & 40 \\
Qwen3.5+       & 4.46  & 45.54 & 91 \\
Qwen3.5-397B   & 25.60 & 24.40 & 49 \\
\midrule
\multicolumn{4}{c}{\textit{Image-based VLM}} \\
\midrule
Gemini-3.1-Pro & 5.15  & 44.85 & 90 \\
Gemini-3-Flash & 1.77  & 48.23 & 96 \\
Seed-1.6       & 6.35  & 43.65 & 87 \\
Seed-1.6-Flash & 30.00 & 20.00 & 40 \\
Qwen3.5+       & 3.48  & 46.52 & 93 \\
Qwen3.5-397B   & 22.03 & 27.97 & 56 \\
\bottomrule
\end{tabular}
}
\vspace{-6mm}
\end{wraptable}

We evaluate these VLMs under two input paradigms (prompt details in the Appendix). First, we use a \textit{Video-based} approach. Because modern VLMs typically subsample frames to manage context length, this inherent preprocessing disrupts temporal spacing, predictably degrading frame rate perception. To bypass this architectural bottleneck, we introduce an \textit{Image-based} paradigm, unrolling the video into 128 discrete images fed sequentially to preserve the absolute frame count and temporal order.

\begin{figure*}[t]
\centering
\includegraphics[width=0.95\linewidth]{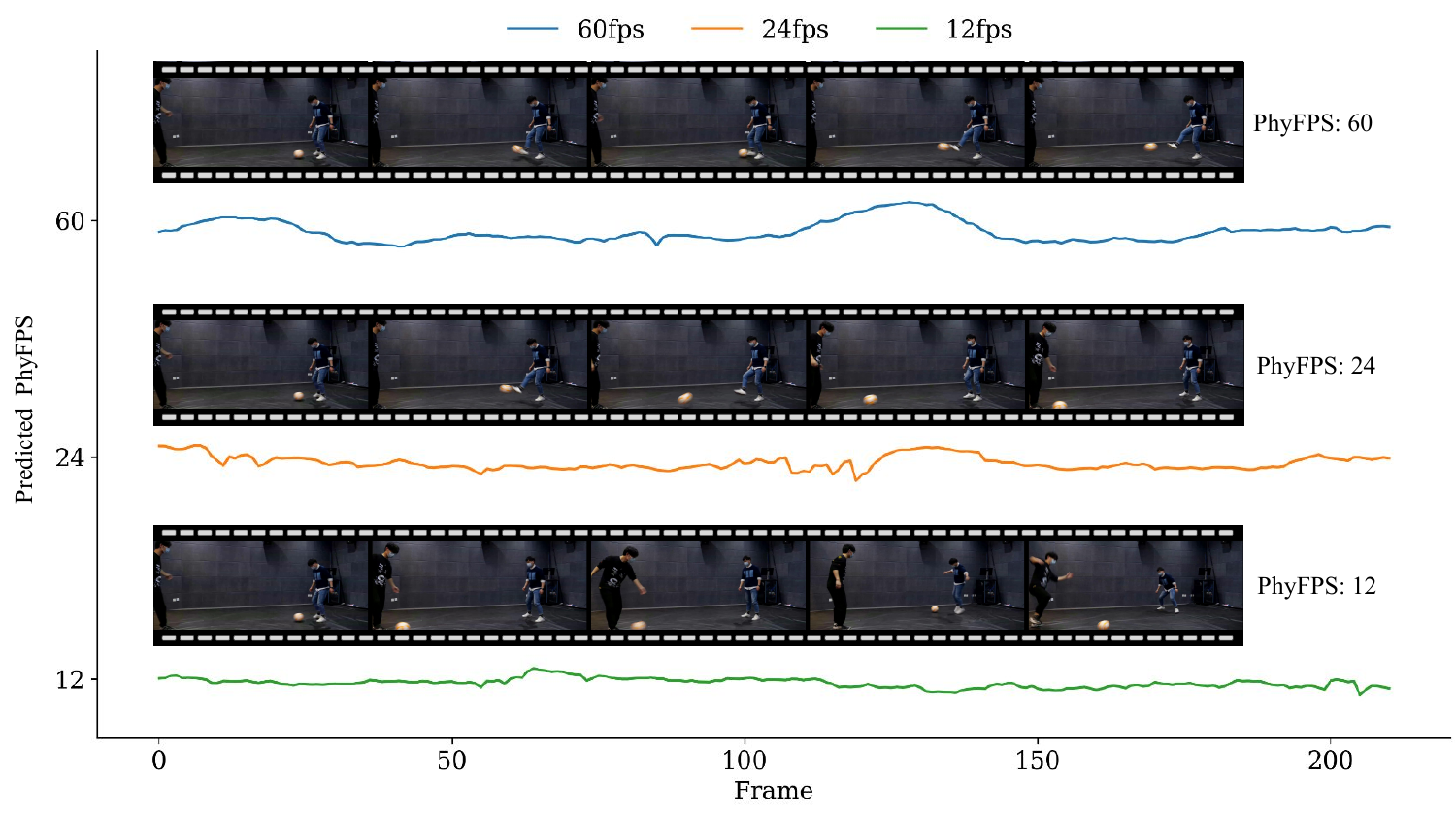}
\vspace{-2mm}
\caption{\textbf{Continuous PhyFPS Prediction on Real Dynamics.} Qualitative results from our Visual Chronometer evaluating a single dynamic action (soccer ball juggling) captured at three distinct physical frame rates (60, 24, and 12 PhyFPS). The model not only accurately recovers the absolute time scale directly from visual cues but also maintains remarkable temporal stability across the entire sequence.}
\label{fig:qualitative_pred}
\end{figure*}

The results (\Cref{tab:fps_comparison}) show that our Visual Chronometers (\textbf{VC-Common} and \textbf{VC-Wide}) achieve exceptionally low MAE and MAPE, with the narrower-range VC-Common predictably yielding the tightest margins. Qualitatively, \Cref{fig:qualitative_pred} confirms our model's ability to continuously and accurately track physical time scales across varying base rates. 

Conversely, all tested VLMs fail catastrophically at physical estimation for both video inputs and unrolled image sequences. Many suffer from severe mode collapse. For example, Seed-1.6-Flash degenerates to predicting exactly 30 FPS for all inputs regardless of the actual dynamics. These findings demonstrate that general-purpose foundation models lack a grounded internal motion pulse, reinforcing the necessity of our specialized architecture.

\subsection{Ablation Studies}
\label{sec:ablation}

To validate our core design choices, we conduct ablation studies on the \textbf{VC-Common} model, evaluating the impact of physics-grounded data augmentations and inference temporal context length.

\begin{table}[h]
\centering
\small
\setlength{\tabcolsep}{10pt}
\caption{\textbf{Ablation Study on Temporal Data Augmentations.} Evaluated on \texttt{PhyFPS-Bench-Real} using the VC-Common configuration.}
\label{tab:ablation_aug}
\begin{tabular}{lcccc}
\toprule
Augmentation Strategy & Motion Blur & Rolling Shutter & MAE $\downarrow$ & MAPE (\%) $\downarrow$ \\
\midrule
Naive Baseline & \ding{55} & \ding{55} & 5.12 & 13 \\
+ Motion Blur & \ding{51} & \ding{55} & 4.87 & 11 \\
\textbf{VC-Common} & \ding{51} & \ding{51} & \textbf{3.46} & \textbf{9} \\
\bottomrule
\end{tabular}
\end{table}

\paragraph{Impact of Temporal Augmentations.} 
To verify the necessity of our physics-grounded augmentations (Fast Shutter, Motion Blur, and Synthetic Rolling Shutter), we train a naive baseline using only uniform temporal subsampling. Evaluated on the in-the-wild conditions of \texttt{PhyFPS-Bench-Real} (\Cref{tab:ablation_aug}), the baseline degrades significantly, with MAE increasing from 3.46 to 5.12. Without simulating exposure integration or sequential sensor readout during training, the naive model overfits to idealized spatial displacements and fails to disentangle physical speed from realistic motion artifacts. This confirms our augmentations are critical for learning robust, intrinsic visual dynamics.

\paragraph{Impact of Temporal Context Length.}
Measuring physical speed computationally requires sufficient kinematic history. We evaluate the robustness of \textbf{VC-Common} across varying inference window lengths (patch sizes) $T \in \{8, 16, 32, 64, 128\}$. 

\begin{figure}[t]
    \centering
    \includegraphics[width=0.95\linewidth]{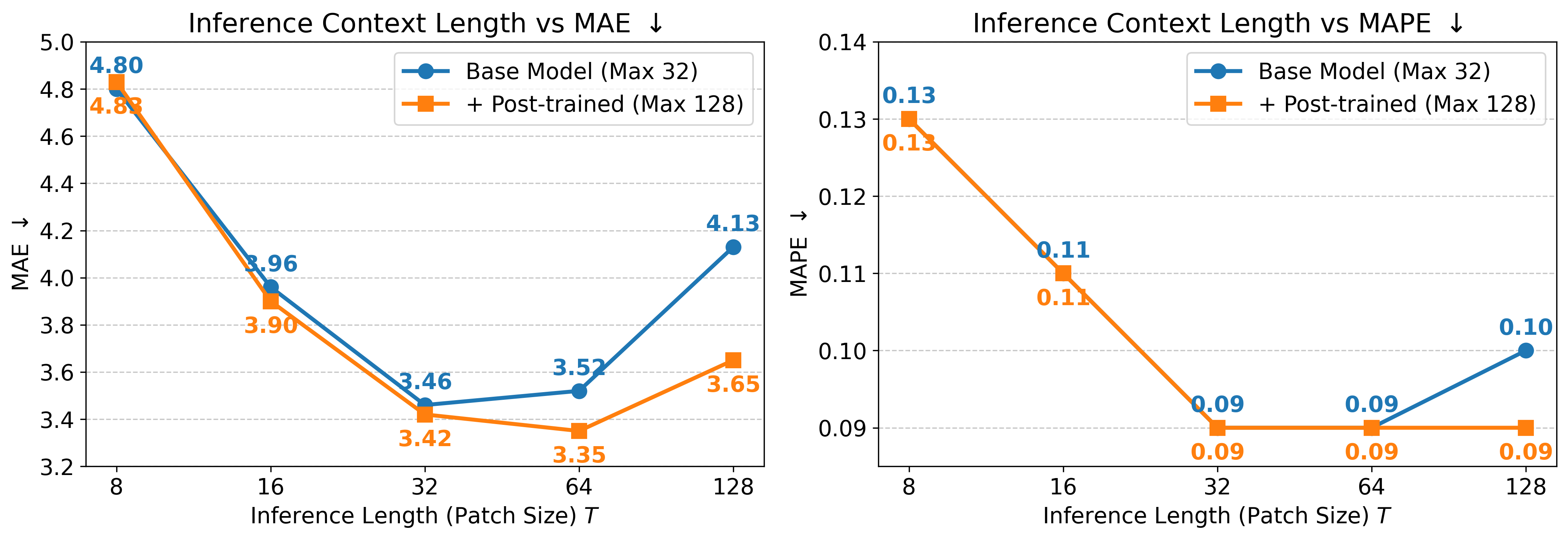}
    \caption{\textbf{Ablation on Inference Context Length ($T$).} Evaluating the VC-Common model across different inference patch sizes on \texttt{PhyFPS-Bench-Real}. We compare the base model (trained on max 32 frames) with a post-trained variant (max 128 frames) to analyze the trade-off between temporal receptive field and sliding-window granularity.}
    \vspace{-4mm}
    \label{tab:ablation_length}
\end{figure}

As illustrated in \Cref{tab:ablation_length}, the base model (trained on max 32 frames) expectedly struggles with ultra-short contexts ($T=8$) due to insufficient visual evidence, optimizing at $T=32$ ($\text{MAE} = 3.46$). Notably, it demonstrates strong length extrapolation, maintaining competitive accuracy at $T=64$ and $128$. Post-training the model on a maximum length of 128 frames further improves performance at $T=64$ without degrading short-patch accuracy. 

However, a critical bottleneck emerges: increasing the inference patch size to $T=128$ fails to outperform $T=64$. This reveals an inherent trade-off in temporal modeling. While small patches lack sufficient receptive fields, extremely large patches (e.g., $T=128$, spanning the entire benchmark video) restrict evaluation to a single global inference pass. This loses the variance-reduction benefits of sliding-window ensembling and strips the model of its ability to capture fine-grained PhyFPS fluctuations within a single shot. Consequently, a mid-range patch size ($T=32$ to $64$) optimally balances kinematic context with local temporal granularity.

%% file: sec/con.tex
\section{Discussion: Implications and Future Directions}
\label{sec:discussion}

In this section, we contextualize our findings and explore future directions for temporal modeling in generative video through a question-and-answer format.

\vspace{2mm}
\noindent
\textbf{Q1: Is strict alignment between PhyFPS and meta FPS always desirable? In other words, is chronometric hallucination inherently problematic, given that intentional speed manipulation is a core creative tool in filmmaking?}

\noindent
\textbf{A:} Dynamic retiming---such as deliberate slow-motion or time-lapse---is undeniably a vital creative tool. We do not argue that every generated video must strictly adhere to a $1\times$ physical time scale (i.e., $\text{PhyFPS} = \text{meta FPS}$). Rather, the fundamental issue with chronometric hallucination lies in the absence of \textbf{controllability}. Currently, models hallucinate time scales arbitrarily; a user prompting for a ``person walking'' might implicitly receive a sequence operating at $0.5\times$ or $2\times$ physical speed without any explicit instruction. While variable speeds are essential for specific creative scenarios, the ability to stably generate a grounded, default $1\times$ speed is a prerequisite for true controllability. If generative video models are to evolve into reliable world models, they must possess a stable internal pulse. Only by first mastering baseline physical reality can a model faithfully execute deliberate $N\times$ speed manipulations upon request.

\vspace{2mm}
\noindent
\textbf{Q2: How can future video generation pipelines resolve chronometric hallucination?}

\noindent
\textbf{A:} To resolve this issue, future pipelines should treat time as an active, controllable condition. First, at the data curation level, training datasets should be rigorously relabeled with their true intrinsic PhyFPS. Our Visual Chronometer can serve as an automated, large-scale annotator to explicitly filter or condition the input distribution. Second, at the architectural level, models require temporal conditioning mechanisms that force the network to explicitly comprehend and disentangle the true pulse of varying physical frame rates during training. Finally, from an optimization standpoint, the Visual Chronometer has the potential to act as a specialized reward model. By providing direct, physics-grounded supervision signals during preference alignment (e.g., via RLHF or DPO), it can guide generative models to strictly adhere to desired temporal dynamics and structurally eliminate chronometric hallucination.

\section{Conclusion}
\label{sec:conclusion}

In this work, we identify and formalize the phenomenon of \textbf{chronometric hallucination} in modern video generative models, where a reliance on arbitrary metadata containers leads to ambiguous and uncontrollable physical speeds. To address this issue, we propose the \textbf{Visual Chronometer}, a robust predictor trained via physics-grounded temporal resampling that accurately recovers the intrinsic Physical Frames Per Second (PhyFPS) directly from visual dynamics. Through our comprehensive benchmarks, \texttt{PhyFPS-Bench-Gen} and \texttt{PhyFPS-Bench-Real}, we reveal a stark reality: state-of-the-art generators and vision-language models currently struggle to maintain a consistent internal pulse of motion. Nevertheless, by demonstrating that PhyFPS-guided dynamic retiming significantly improves the human-perceived temporal naturalness of AI-generated videos, we offer an immediate, practical mitigation. Ultimately, we hope this work inspires future generative world models to transition from passive metadata reliance to active, physics-grounded temporal conditioning.